\documentclass[manuscript,screen,nonacm]{acmart}

\usepackage{booktabs}
\usepackage{multirow}
\usepackage{arydshln}
\usepackage{algorithm}
\usepackage{algorithmic}

\usepackage{blindtext}

\usepackage{xpatch}

\makeatletter
\xpatchcmd{\ps@firstpagestyle}{Manuscript submitted to ACM}{}{\typeout{First patch succeeded}}{\typeout{first patch failed}}
\xpatchcmd{\ps@standardpagestyle}{Manuscript submitted to ACM}{}{\typeout{Second patch succeeded}}{\typeout{Second patch failed}}    \@ACM@manuscriptfalse
\makeatother

\renewcommand\footnotetextcopyrightpermission[1]{} 
\setcopyright{none}
\AtBeginDocument{%
  \providecommand\BibTeX{{%
    \normalfont B\kern-0.5em{\scshape i\kern-0.25em b}\kern-0.8em\TeX}}}

\setcopyright{none}

\acmConference[arxiv]{Arxiv}

\begin{document}

\title{Dependency and Span, Cross-Style Semantic Role Labeling on PropBank and NomBank}

\author{Zuchao Li}
\email{charlee@sjtu.edu.cn}

\author{Hai Zhao}
\authornote{Corresponding author.}
\email{zhaohai@cs.sjtu.edu.cn}

\author{Junru Zhou}
\email{zhoujunru@sjtu.edu.cn}

\author{Kevin Parnow}
\email{parnow@cs.sjtu.edu.cn}

\author{Shexia He}
\email{heshexia@sjtu.edu.cn}

\affiliation{%
  \institution{Department of Computer Science and Engineering; Key Laboratory of Shanghai Education Commission for Intelligent Interaction and Cognitive Engineering; MoE Key Lab of Artificial Intelligence, AI Institute, Shanghai Jiao Tong University}
  \streetaddress{SEIEE Building \#03-417}
  \city{Shanghai}
  \country{China}
  \postcode{200240}
}

\renewcommand{\shortauthors}{Li et al.}

\begin{abstract}
  The latest developments in neural semantic role labeling (SRL) have shown great performance improvements with both the dependency and span formalisms/styles. Although the two styles share many similarities in linguistic meaning and computation, most previous studies focus on a single style. In this paper, we define a new cross-style semantic role label convention and propose a new cross-style joint optimization model designed around the most basic linguistic meaning of a semantic role, providing a solution to make the results of the two styles more comparable and allowing both formalisms of SRL to benefit from their natural connections in both linguistics and computation. Our model learns a general semantic argument structure and is capable of outputting in either style. Additionally, we propose a syntax-aided method to uniformly enhance the learning of both dependency and span representations. Experiments show that the proposed methods are effective on both span and dependency SRL benchmarks.
\end{abstract}



\maketitle

\section{Introduction}

Semantic role labeling (SRL) aims to derive linguistic meaning representations, such as the predicate-argument structure, for a given sentence. The currently popular formalisms for representing the semantic predicate-argument structure are dependency-based or span-based. While dependency SRL annotates the syntactic heads of arguments, the span style annotates entire argument spans.

Both the dependency and span styles are effective formal representations for semantics, but which is superior has been uncertain for a long time. Furthermore, researchers have suspected that these two SRL models may benefit from being developed together rather than separately. This topic has been roughly discussed by \citet{johansson2008dependency} and \citet{li2019dependency}, who both concluded that the best dependency SRL system at the time clearly outperformed the best span-based system through gold syntactic structure transformation. Additionally, \citet{peng2018learning} integrated dependency and span style SRL into a model for frame-semantic parsing using a multi-task learning approach, which enabled the model to learn the internal relationship between dependency and span styles from multiple datasets. 

In general, current research is limited to the existing datasets and argument representations, which adopt constituent-to-dependency head-rule transformations or multi-task joint learning methods to make the results of dependency and span SRL more comparable. Therefore, in this work, we create a new SRL dataset, ConSD, with a more general style of argument structure. Additionally, we take a new argument structure formulization that enables our model to use a single decoder to implement both semantic formalisms. To verify the effectiveness and applicability of the proposed method, we evaluate the model on the ConSD, CoNLL-2005 (CoNLL-05), and  CoNLL-2009  (CoNLL-09)  benchmarks and cross-evaluate ConSD trained model with both CoNLL benchmarks. Experimental results, therefore, show that the proposed general argument structure is effective for both main SRL formalisms and the proposed dataset is complete.

\section{Dataset}\label{sec:dataset}

Two styles of semantic role labeling have emerged since the PropBank \cite{propbank} style semantic annotation was applied to Pen Treebank  (PTB) \cite{marcus1993building}: span-based and dependency-based. Early semantic role labeling (CoNLL-2004 and CoNLL-2005 shared tasks) on the Penn Treebank was span-based \cite{carreras2004introduction,CoNLL2005} with spans corresponding to syntactic constituents; however, as in syntactic parsing, there are sometimes theoretical or practical reasons to prefer dependency graphs. To this end, the CoNLL-2008 shared task \cite{surdeanu-EtAl2008} proposed a unified dependency-based formalism that models both syntactic dependencies and semantic roles in addition to introducing the nominal predicate-argument structure from NomBank \cite{nombank}. The CoNLL-2009 shared task \cite{hajivc-EtAl2009} further built on the CoNLL-2008 shared task by providing six more language annotations in addition to the original English. CoNLL-2005 and CoNLL-2009 have been the benchmark datasets for span-based and dependency-based SRL, respectively. Although both CoNLL-2005 and CoNLL-2009 followed PropBank, due to the different labeling, conversion times, and standards, as well as the use of NomBank in the CoNLL-2009, there are large differences in the data of the two tasks. Therefore, there exists a bifurcation in semantic role labeling research preventing the results from being directly compared, let alone from enjoying the benefits from a method of joint semantic learning.

We seek to reduce the style-specific balkanization in the field of semantic role labeling. Our goals, therefore, include (a) a unifying formal model over different-style semantic role labeling treebanks, (b) uniform representations and scoring, (c) systematic contrastive evaluation across styles, and (d) increased cross-fertilization via transfer and multi-task learning. We hope to uniformize the representations of different styles, including those from the two types of prior style-specific semantic role labeling tasks at CoNLL-2004, CoNLL-2005, CoNLL-2008, and CoNLL-2009.  Owing to the scarcity of semantic annotations across different styles, the shared tasks are regrettably limited to parsing English on PTB text for the time being. For the first time, this work combines formallistically and linguistically different approaches to semantic role labeling representation in one uniform form with a single training and evaluation setup. We aim to develop parsing systems that support two distinct semantic role labeling styles in a single model, which encode all core predicate-argument structure. Learning from multiple styles of semantic role labeling representation in tandem has seldom been explored (with notable exceptions, e.g., the parser of  \cite{peng2018learning} on FrameNet).

In order to achieve the above goals, we generate a Consistent Span and Dependency dataset (\textbf{ConSD}) through a process that merges several source treebanks and converts them from the constituent-based and dependency-based formalisms to a new uniform formalism. In this section, we will introduce our uniform format, the source treebanks used, and the conversion process.

\subsection{Uniform Format}

For span-based SRL, the format of the predicate-argument structure is a quadruple consisting of the predicate, the start of argument span, the end of argument span, and the corresponding predicate-argument relation. For dependency SRL, the structure is a triple consisting of the predicate, the syntactic head of the argument span, and their relationship. To handle both representations uniformly, we design a uniform format to represent the two styles, which is a quintuple consisting of the predicate, the start of argument span, the end of argument span, the syntactic head of argument span, and the predicate-argument relation.

For example, given an input text \emph{The bill would have lifted the minimum wage of working to \$ 4.55 an hour by late 1991}, one of the predicates is \textit{lifted}. In span-based SRL, the \textbf{ARG1} argument is [\textit{\textbf{the} minimum wage of \textbf{working}}], while in dependency-based SRL, the argument is (\textit{\textbf{wage}}), which is the dependency head of the argument span in span-based SRL.

\begin{figure}
	\begin{minipage}{0.45\linewidth}
		\begin{align*}
		\textbf{S:} & \quad \textit{lifted}_{\textbf{PRED}} \quad [\textit{\textbf{the} minimum wage} 
		\\ & \quad\quad\quad\quad\quad\quad\quad\quad\quad \textit{of \textbf{working}}]_{\textbf{ARG1}}, \\
		\textbf{D:} & \quad \textit{lifted}_{\textbf{PRED}} \quad (\textit{\textbf{wage}})_{\textbf{ARG1}}, \\
		\textbf{U:} & \quad \textit{lifted}_{\textbf{PRED}} \quad [\textit{\textbf{the} minimum wage}
		\\ & \quad\quad\quad\quad\quad\quad\quad\quad\quad \textit{of \textbf{working}}](\textit{\textbf{wage}})_{\textbf{ARG1}}.
		\end{align*}
	\end{minipage}
	\hfill
	\begin{minipage}{0.49\linewidth}
		\centering
		\includegraphics[width=0.9\textwidth]{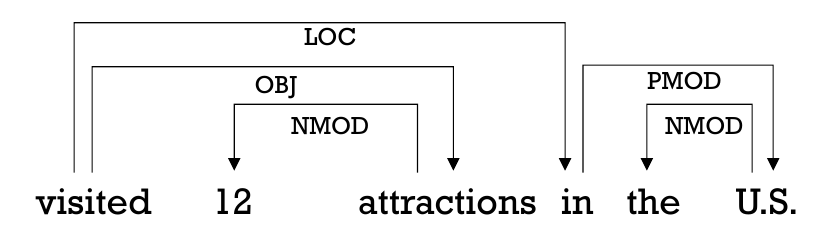}
		\caption{Partial dependency tree for text \emph{visited 12 attractions in the U.S.}.}
		\label{fig:partial_syntax}
	\end{minipage}
\end{figure}

We combine the two formalisms, span-based (\textbf{S}) and dependency-based (\textbf{D}), to create a new uniform SRL format (\textbf{U}) to predict both span-based and dependency-based arguments at the same time. This combination has several linguistic benefits. Unlike span-based SRL, we do not only focus on the boundary of the argument; we also focus on the center of the argument (the dependency head). Unlike dependency-based SRL, not only has the center been attended to, but the range of constituent in the sentence has also been considered. For the application of downstream tasks, this uniform format can extract the argument range and core components without the help of syntactic parsing. Compared to using syntactic parsing, this is a more conducive strategy, as it has less potential error-propagation.

\subsection{Source Treebank and Banks}

Similar to the merging procedures in the CoNLL-2008 shared task, source treebanks for our conversion process also include the PTB, PropBank, and NomBank.

\paragraph{Penn Treebank 3 (PTB)}
The Penn Treebank 3 \cite{marcus1993building} consists of hand-coded parses of the Wall Street Journal (training, development, and test) and a small subset of the Brown corpus \cite{francis1979brown} (out-of-domain, test only). The Penn Treebank syntactic annotation includes phrases, part-of-speech (POS), and empty category representations of various filler/gap constructions and other phenomena and is based on a theoretical perspective similar to that of Government and Binding Theory \cite{chomsky1993lectures}. We follow the standard partition used in syntactic parsing: sections 02-21 for training, section 24 for development, and section 23 for test. In addition, three sections (ck01-03) from the Brown corpus are used for out-of-domain evaluation setting. 

\paragraph{Proposition Bank I (PropBank)}
The PropBank annotation \cite{propbank} classifies the arguments of all the main verbs in PTB. Arguments are numbered (ARG0, ARG1, ...) based on lexical entries or frames. Different sets of arguments are assumed for different rolesets. Dependent constituents that fall into categories independent of the lexical entries are classified as various types of ARGM (TMP, ADV, etc.). Rather than using PropBank directly, we used the version created for the CoNLL-2005 shared task \cite{CoNLL2005}.

\paragraph{NomBank}

The NomBank annotation \cite{nombank} essentially uses the same framework as PropBank to annotate arguments of nouns. Differences between PropBank and NomBank stem from (1) differences between noun and verb argument structure, (2) differences in treatment of nouns and verbs in PTB, and (3) differences in the sophistication of previous research about noun and verb argument structure.

\subsection{Processing Steps}

\paragraph{Re-tokenization} 
Since NomBank uses a subword analysis in some hyphenated words (such as
$[\textit{finger}]_{\textbf{ARG}}$-$[\textit{pointing}]_{\textbf{PRED}}$), we need to re-tokenize the words provided in the original PTB. Specifically, we split the Treebank tokens at a hyphen (\textit{-}) or a forward slash (\textit{/}) if the segments on either side of these delimiters are: (a) a word in a dictionary (COMLEX Syntax or any of the dictionaries available on the NOMLEX website); (b) part of a markable Named Entity; or (c) a prefix from the list: \textit{co}, \textit{pre}, \textit{post}, \textit{un}, \textit{anti}, \textit{ante}, \textit{ex}, \textit{extra}, \textit{fore}, \textit{non}, \textit{over}, \textit{pro}, \textit{re}, \textit{super}, \textit{sub}, \textit{tri}, \textit{bi}, \textit{uni}, \textit{ultra}. For example, \textit{McGraw-Hill} was split into 3 segments: \textit{McGraw}, \textit{-}, and \textit{Hill}. This step is consistent with the CoNLL-2008 shared task.

\paragraph{Lemmas and Parts-of-Speech} 
For ease of use, refererencing the CoNLL-2009 shared task data, we provide the same gold-standard lemmas, automatically predicted lemmas,  gold-standard POS tags, and automatically predicted POS tags.

\paragraph{Constituent Syntactic Trees}
The golden syntactic trees are provided for the PTB; however, due to the re-tokenization, we need to adjust the spans of the constituents. For the predicted syntactic tree, we use the paser of \cite{Kitaev-2018-SelfAttentive} to obtain full-parses.

\paragraph{Dependency Syntactic Tree}

Since the dependency syntax trees are not provided on PTB, we thus have to convert the dependency trees automatically from the PTB. The dependency syntax represents grammatical structure by means of labeled binary head-dependent relations rather than phrases. The idea underpinning constituent-to-dependency conversion algorithms \cite{magerman1994natural,collins2003head,yamada2003statistical,de2006generating} is that head-dependent pairs are created from constituents by selecting one word in each phrase as the head and setting all others as its dependents. The dependency labels are then inferred from the phrase-subphrase or phrase-word relations. There are three typical different conversion rules and tools on PTB: (1) Penn2Malt and the head rules of \cite{yamada2003statistical} , noted as PTB-Y\&M, (2) the dependency converter in the Stanford Parser v3.3.0 with Stanford Basic Dependencies \cite{de2006generating}, noted as PTB-SD, and (3) the LTH Constituent-to-Dependency Conversion Tool \cite{johansson2007extended}, noted as PTB-LTH. In order to make the results on our corpora as comparable as possible to those using the CoNLL-2009 shared task, we use the same PTB-LTH conversion method to convert the golden and predicted constituent trees into dependency trees.

\paragraph{Predicate}
We transform all annotated semantic arguments (not just a subset) in PropBank and NomBank and address propositions centered around both verbal (PropBank) and nominal (NomBank) predicates. 

\paragraph{Uniform Argument}

In order to obtain the uniform argument representation, the first step is to convert the underlying constituent structure of PropBank and NomBank in a manner similar to those of CoNLL-2008 and CoNLL-2009. Just as in syntax constituent-to-dependency conversion algorithms, finding the dependency argument is akin to finding the head of the span on the dependency tree. In order to avoid replicating efforts and to ensure compatibility between syntactic and semantic dependencies, we use only the span argument boundaries and the golden dependency tree. We identify the head of an argument span using the following heuristic: 

\emph{The head of a semantic argument is assigned to the token inside the argument boundaries whose dependency head is a token outside the argument boundaries} \cite{surdeanu-EtAl2008}.

For example, consider the following annotated text: [\textit{visited}]$_{\textbf{PRED}}$ [\textit{12 attractions}]$_{\textbf{ARG1}}$ [\textit{in the U.S.}]$_{\textbf{ARGM-LOC}}$, whose syntax tree is shown in Figure \ref{fig:partial_syntax}. According to the heuristic, the head of $\textbf{ARG1}$ argument is set to \textit{attractions}, as the dependency head of \textit{attractions} is \textit{visited}, which is outside of span $[\textit{12 attractions}]$. Similarly, the head of $\textbf{ARGM-LOC}$ is set to \textit{in}.

\begin{figure}
	\begin{minipage}{0.49\linewidth}
		\centering
		\includegraphics[width=1.0\textwidth]{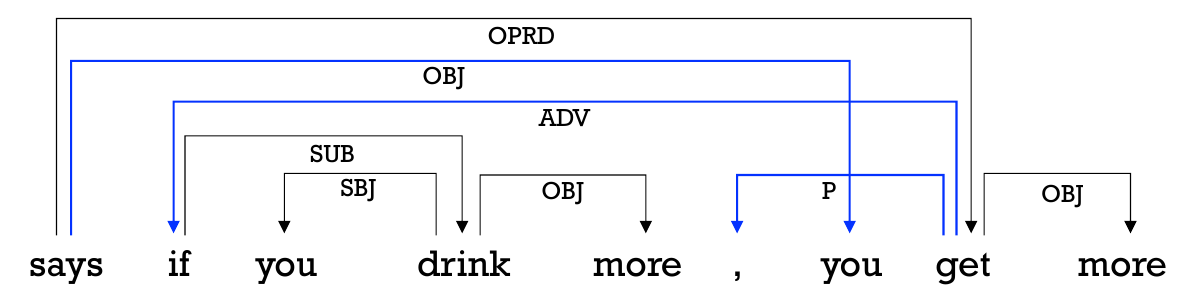}
		\caption{Partial non-projective dependency tree for text \emph{says if you drink more, you get more}.}
		\label{fig:non_projective}
	\end{minipage}
	\hfill
	\begin{minipage}{0.49\linewidth}
		\centering
		\includegraphics[width=1.0\textwidth]{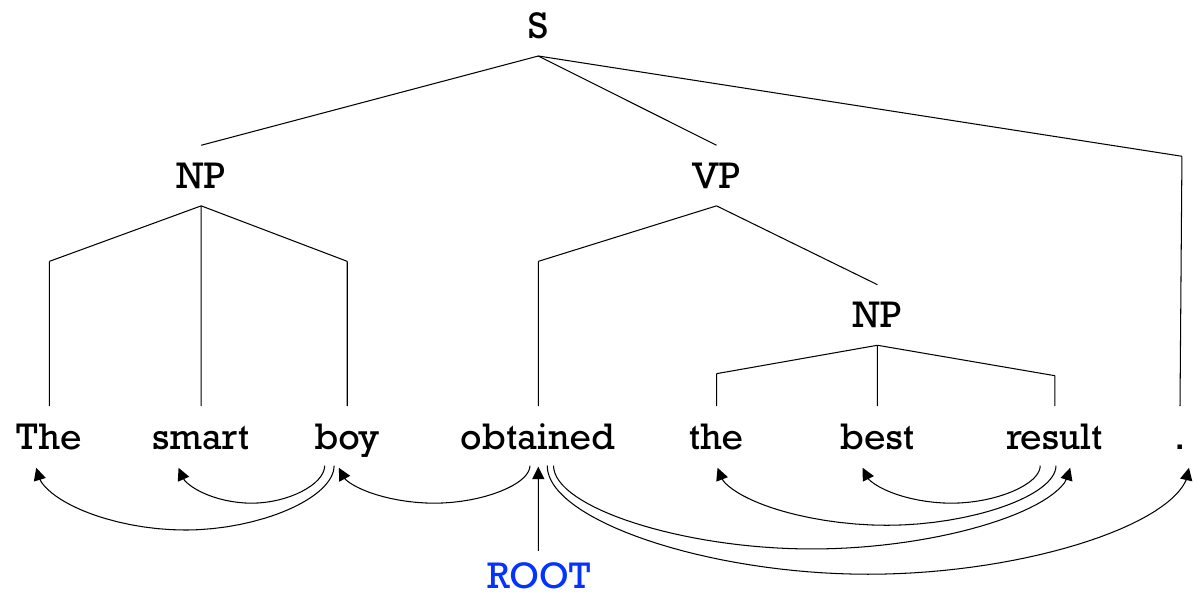}
		\caption{An example of structures of constituency syntax tree (above the sentence) and dependency syntax tree (below the sentence). The POS tags of the constituency tree and the labels of the dependency tree are omitted for simplicity.}
		\label{fig:syntax_tree}
	\end{minipage}
\end{figure}

While the heuristic works well on the majority of arguments and guarantees a one-to-one relationship between the span argument and the head, some dependency trees have special structures, and some span arguments have multiple heads. Both these situations affect the aforementioned one-to-one relationship, so we need to deal with these cases individually.

For 0.7\% of the span arguments, multiple syntactic heads are detected inside the span boundary. For example, [\textit{allow}]$_{\textbf{PRED}}$ [\textit{executives to report exercises of options later and less often}]$_{\textbf{ARG1}}$. Under our rule set defined by our processing script\footnote{Note this rule set is the counterpart of syntactic constituent-to-dependency conversion, which is usually defined by "head rules."}, two syntactic heads \textit{executives} and \textit{to} are assigned \textbf{ARG1}. Therefore, we split the original span argument into a sequence of discontinuous sub-spans and set the first sub-span to the original argument role $role$ and the rest to C-$role$ - the \textbf{ARG1} argument becomes [\textit{allow}]$_{\textbf{PRED}}$ [\textit{executives}]$_{\textbf{ARG1}}$ [\textit{to report exercises of options later and less often}]$_{\textbf{C-ARG1}}$, for example.

In non-projective dependency trees, all the children of a head cannot form a continuous span, so it is necessary to continue iterative splitting of the discontinuous span until all sub-spans have only one head and all children of each head compose a continuous span. For example, [\textit{says}]$_{\textbf{PRED}}$ [\textit{if you drink more, you get more}]$_{\textbf{ARG1}}$. In the first iteration, two syntactic heads \textit{you} and \textit{get} are detected, and the original span argument is split into two sub-spans [\textit{you}]$_{\textbf{ARG1}}$ [\textit{if you drink more, get more}]$_{\textbf{C-ARG1}}$. Since the second sub-span [\textit{if you drink more, get more}]$_{\textbf{C-ARG1}}$ is not continuous, it cannot be used as a constituent. Therefore, further splitting is required until each span is continuous and has a unique syntax head. We use the recursive partitioning method and finally get the span as follows: [\textit{if you drink more}]$_{\textbf{ARG1}}$ (\textit{if}), [\textit{,}]$_{\textbf{C-ARG1}}$ (\textit{,}), [\textit{you}]$_{\textbf{C-ARG1}}$ (\textit{you}), and [\textit{get more}]$_{\textbf{C-ARG1}}$ (\textit{get}).

\subsection{Dataset Comparison}

We analyzed the number of sentences, words, predicates, and arguments in different splits of the original CoNLL-05, CoNLL-09 as well as our ConSD datasets. The statistical results are listed in Table \ref{tab:data_stat}.
Comparing the statistics of CoNLL-05 and ConSD, we can see that the number of sentences in ConSid and CoNLL-05 remained the same, but the number of words increased, resulting from the re-tokenization process. The number of predicates also increased by nearly 50\% due to the introduction of nominal predicates in NomBank. Accompanying the introduction of nominal predicates, the arguments also increased. Comparing CoNLL-09 to ConSD, since CoNLL-09 drops some sentences present in CoNLL-05, in order to make our dataset ConSD to cover both CoNLL-05 and CoNLL-09, we restore the sentences and manually annotate  them. Changes in the number of words, predicates, and arguments are thus due to these additional sentences. In general, our dataset ConSD is the union of the existing CoNLL-05 and CoNLL-09 datasets and provides a more complete span SRL annotation (containing nominal predicates) compared to CoNLL-05.

\begin{table}[h]
	\centering
	\small
	\setlength{\tabcolsep}{4pt}
	\caption{Data statistics of sentences, words, predicates, and arguments in different splits from the original CoNLL-05 and CoNLL-09 and our ConSD Treebank.}\label{tab:data_stat}
	\begin{tabular}{lcccccccccccc}
		\toprule  
		&\multicolumn{4}{c}{Original CoNLL-05}&\multicolumn{4}{c}{Original CoNLL-09}& \multicolumn{4}{c}{\bf ConSD Dataset}\\  
		\cmidrule(lr){2-5} \cmidrule(lr){6-9}\cmidrule(lr){10-13}
		&Train&Dev&WSJ&Brown&Train&Dev&WSJ&Brown& Train&Dev &WSJ&Brown\\  
		\midrule
		\bf \#SENTS & 39,832 & 1,346 & 2,416 & 426 & 39,279 & 1,334 & 2,399 & 425 & 39,832 & 1,346 & 2,416 & 426 \\
		\bf \#WORDS & 95,0028 & 32,853 & 56,684 & 7,159 & 958,167 & 33,368 & 57,676 & 7,207 & 974,784 & 33,763 & 58,226 & 7,235 \\
		\bf \#PREDS & 90,750 & 3,248 & 5,267 & 804 & 179,014 & 6,390 & 10,498 & 1,259 & 180,571 & 6,423 & 10,498 & 1,259 \\
		\bf \#ARGS & 245,353 & 8,549 & 14,463 & 2,228 & 393,699 & 13,865 & 23,286 & 2,859 & 399,468 & 13,990 & 23,286  & 2,859\\
		\bottomrule
	\end{tabular}
\end{table}

\section{Baseline}
\begin{figure*}[t]
	\centering
	\includegraphics[scale=0.8]{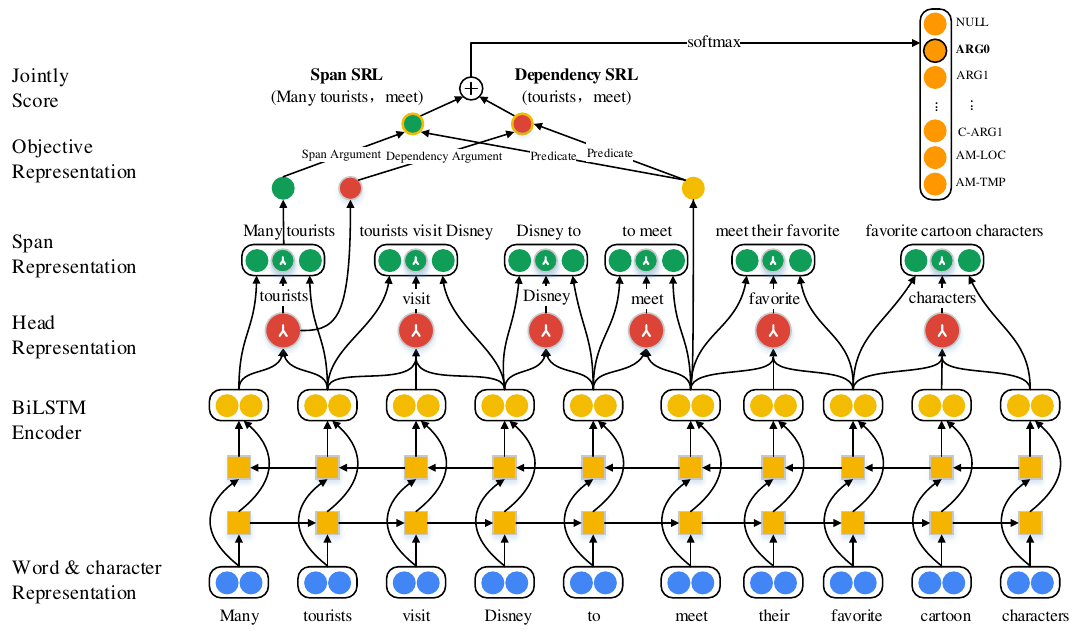}
	\caption{\label{fig:overview} Overall architecture of our proposed SRL model.}
\end{figure*}

To demonstrate the performance on the new ConSD dataset, we set a baseline model and conducted experiments on this baseline model. 
Given a sentence, the SRL task can be decomposed into four classification subtasks: predicate identification, predicate disambiguation, argument identification, and argument classification; however, most previous efforts have focused on argument identification and argument classification, while predicate identification, despite its common neglect, is still very much needed in many downstream applications. Therefore, we propose an end-to-end SRL model to jointly predict all predicates and general arguments and the relations between them. Our model is built on one of the recent successful syntax-agnostic SRL models \cite{li2019dependency}. As shown in Figure \ref{fig:overview}, our model handles argument identification and classification in one step by treating the SRL task as predicate-argument pair classification and consists of four main modules: (1) a bidirectional LSTM (BiLSTM) encoder, (2) a general representation layer that takes as input the BiLSTM representations, (3) a syntax aid component, and (4) a biaffine scorer that takes the predicate and its argument candidates as input.

\subsection{BiLSTM encoder}
Given a sentence $S = \{w_1,...,w_n\}$, we adopt an $m$-layer bidirectional LSTM (BiLSTM) with highway connections (denoting it as HBiLSTM) to obtain contextualized representations, which takes as input the token representation. Following \cite{he2018jointly}, the token representation $e_i$ consists of pre-trained word embeddings $e^{word}$ concatenated with character-based representations $e^{char}$. We also further enhance the token representation by concatenating an external pre-trained language model's layer features $e^{lm}$ from the recent successful language models. The final contextualized representation $x_i$ can be denoted as:
\begin{equation}
x_i = \textit{BiLSTM}([e_i^{word};e_i^{char};e_i^{lm}]).
\end{equation}

\subsection{Objective Representation Layer}

Formally, for each candidate argument $a \in \mathcal{A}$, the representation contains four features: the two boundary hidden states of the span argument (start position $\text{S}\text{\tiny{TART}}(\cdot)$ and end position $\text{E}\text{\tiny{ND}}(\cdot)$) from the HBiLSTM outputs $(x_{\text{S}\text{\tiny{TART}}(a)}, x_{\text{E}\text{\tiny{ND}}(a)})$, the attention-based span representation $x_{span}$, and the embedded span width features $e^{width}$. The argument representation $x_{arg}$ can thus be described as follows:
\begin{equation}\label{eq:span_repr}
x_{arg} = [x_{\text{S}\text{\tiny{TART}}(a)}; x_{\text{E}\text{\tiny{ND}}(a)}; x_{span}; e^{width}],
\end{equation}
\begin{equation}\label{eq:span_attn}
\alpha_{span} = \textbf{softmax}(\textbf{w}^T_s x_{\text{S}\text{\tiny{TART}}(a):\text{E}\text{\tiny{ND}}(a)}),
\end{equation}
\begin{equation}
x_{span} = x_{\text{S}\text{\tiny{TART}}(a):\text{E}\text{\tiny{ND}}(a)} \cdot \alpha_{span}
\end{equation}
where $x_{\text{S}\text{\tiny{TART}}(a):\text{E}\text{\tiny{ND}}(a)}$ is shorthand for stacking a list of vectors $x_t$ ($\text{S}\text{\tiny{TART}}(a) \leq t \leq \text{E}\text{\tiny{ND}}(a)$), and $\alpha_{span}$ is the span attention weight. In addition, the predicate representation $x_p$ is simply the HBiLSTM output at the candidate's position $p$. At the same time, $\alpha_{span}$ is not only used to obtain the attention-based representation of the span $x_{span}$; it can also be used to train to obtain the dependency head position $h$ of the span:
\begin{equation}
h = \arg\max(\alpha_{span}).
\end{equation}

\subsection{Candidates Pruning}

The number of all possible candidate arguments for a sentence of length $n$ is $\textit{O}(n^2)$ for span SRL and $\textit{O}(n)$ for dependency SRL; however, when we adopt a unified goal, the number is $\textit{O}(n^2)$. As the model deals with $\textit{O}(n)$ possible predicates, the overall computational complexity is $\textit{O}(n^3)$, which is too computationally expensive and why we perform candidate pruning. Following \cite{he2018jointly}, we adopt two unary scorers to choose the most probable candidate arguments and predicates to reduce the overall number of candidate tuples to  $\textit{O}(n^2)$. Furthermore, for arguments, we set the maximum width of an argument to $\mathcal{L}$, which ensures that the number of candidate arguments is only $\textit{O}(n)$.

\subsection{Dependency Syntax Aid (DSA)}

Dependency syntax provides binary asymmetric relations (e.g., modifications and arguments) between words. It is represented by a tree structure with words as nodes and relations as edges. As seen from the data conversion process, the dependency syntax provides the location information of the head token in the argument span. This therefore motivates us to use the dependency tree structure to aid semantic role labeling. 

To utilize such dependency tree structures, for each candidate span $span = \{w_j, w_{j+1}, ..., w_{j+\mathcal{L}}\}$, we get the dependency syntax head set $headset = \{w_h\}, h \in [j:j+\mathcal{L}]$ from the span by the heuristic defined in previous section. We define an indicator embedding $e^{dsa}$ on the dependency syntax head set $headset_t$ input to the calculate the span representation $x_{span}$ and head position $h$.

\begin{equation}
e^{dsa}_t = \begin{cases}
1, & w_t \in headset\\
0, & w_t \notin headset
\end{cases}
\end{equation}

After we add the indicator embedding $e^{dsa}$ into Eq. (\ref{eq:span_attn}), the equation becomes:
\begin{equation}
\alpha_{span} = \textbf{softmax}(\textbf{w}^T_s [x_{\text{S}\text{\tiny{TART}}(a):\text{E}\text{\tiny{ND}}(a)}; e^{dsa}_{\text{S}\text{\tiny{TART}}(a):\text{E}\text{\tiny{ND}}(a)}]).
\end{equation}

\subsection{Constituency Syntax Aid (CSA)}

Constituency syntax breaks a sentence into constituents (i.e., phrases), which naturally forms a constituency tree in a top-down fashion. In contrast with the dependency syntax tree, words can only be the terminals in a constituency tree, while the non-terminals are phrases with types. In SRL, each argument corresponds to a constituent in constituency trees, which can be used to generate span argument candidates given the predicates \cite{xue2004,CoNLL2005}. \cite{punyakanok2005necessity} showed that the constituency tree offers high-quality argument boundaries. 

In order to utilize such constituent boundaries in the constituency tree and help decide argument candidates, we extract all constituent $c$ boundaries to form a set $boundaryset = \{(\text{S}\text{\tiny{TART}}(c),\text{E}\text{\tiny{ND}}(c))\}$. We also define an indicator embedding $e_{csa}$ on the constituent boundaries set $boundaryset$ input to calculate the span representation.
\begin{equation}
e^{csa}_t = \begin{cases}
1, & span_t \in boundaryset\\
0, & span_t \notin boundaryset
\end{cases}
\end{equation}

After we add the indicator embedding $e^{csa}$ into Eq. (\ref{eq:span_repr}), the equation becomes:
\begin{equation}
x_{arg} = [x_{\text{S}\text{\tiny{TART}}(a)}; x_{\text{E}\text{\tiny{ND}}(a)}; x_{span}; e^{width}; e^{csa}].
\end{equation}

\subsection{Scorer}
As mentioned above, our model treats the SRL task as a predicate-argument pair classification problem, handling argument identification and classification in one-shot. To label semantic roles, we employ a scorer with biaffine attention \cite{dozat2017deep} as a role classifier on top of the objective representation layer to make the final prediction, following \cite{cai2018full}. Taking as input the candidate predicates and arguments representations, denoted by $h_p = x_p$ and $h_a = x_{arg}$ respectively, the biaffine scorer computes the probability of the semantic role using biaffine transformation as follows: 
$$\Phi_{r}(p, a) = \{h_p\}^T \textbf{W}_1 h_a + \textbf{W}_2^T (h_p \oplus h_a) + \textbf{b}.$$
where $\oplus$ represents the concatenation operator, $\textbf{W}_1$ and $\textbf{W}_2$ denote the weight matrix of the bilinear and the linear terms respectively, and $\textbf{b}$ is the bias item.

\subsection{Training Objectives}

Our model is trained to optimize the probability of the semantic roles $y_{p,a}$ for a given sentence $x$. The probability $P_\theta(y_{p,a}|x)$ can be factorized as:
$$P_\theta(Y|X)  = \prod_{p \in \mathcal{P}, a \in \mathcal{A}} P_\theta(y_{p,a}|X),$$
$$P_\theta(y_{p,a} = r|X) = \frac{\exp(\Phi_{r}(p, a))}{\sum_{r' \in \mathcal{R}}\exp(\Phi_{r'}(p, a))},$$
where $\theta$ represents learnable model parameters. Thus, for each input X, our model minimizes the negative log likelihood of the span $Y_{span}^*$ and dependency $Y_{dep}^*$ gold structure:
\begin{align*}
\mathcal{J}(X) = &\lambda (-\log P_\theta(Y_{span}^* |X)) \\ + &(1 - \lambda) (-\log P_\theta(Y_{dep}^* |X)),
\end{align*}
where $\lambda$ balances the two formalisms learning during training.

\section{Experiments}
We experimented on our converted dataset ConSD, which was derived from PTB, PropBank, and NomBank, and we use the two common evaluation setups: \emph{end-to-end} and \emph{gold predicates}. The model was evaluated on the micro-averaged $F_1$ for correctly predicting tuples (predicate, argument span, argument head, and label). For the predicate disambiguation task in dependency SRL, following \cite{marcheggianiEMNLP2017}, we used the off-the-shelf disambiguator from \cite{roth2016} and reported the Sem-$F_1$ score using CoNLL-09 official scorer.

\subsection{Setup}
In our experiments, the pre-trained word embeddings are 100-dimensional GloVe vectors \cite{penningtonEMNLP2014}. 
The dimension of the ELMo \cite{ELMo} or BERT \cite{bert2018} language model feature embedding is 1024. Additionally, we used a 3 layer BiLSTM with 400-dimensional hidden states, applying dropout with an 80\% keep probability between time-steps and layers. For the biaffine scorer, we employed two 300-dimensional affine transformations with ReLU non-linear activation and set the dropout probability to 0.2. 
All models were trained for up to 500 epochs with batch size 64 and Adam optimizer initial learning rate $2e^{-3}$. The value of $\lambda$ in our joint loss was set to 0.5.
In this paper, \textbf{Baseline} is equivalent to \citet{li2019dependency}'s model without cross-style joint training, while \textbf{Baseline-joint} denotes our proposed model with joint training between dependency and span annotations. \textbf{+Predicted/Gold Syntax} represents the models that use DSA and CSA for predicted or gold syntax integration based on joint \textbf{Baseline-joint} model. \textbf{Full Model} refers to the model using predicted syntax.

\begin{table*}
	\centering
	\small
	\caption{SRL results on the ConSD test sets with \emph{gold predicates} setup. The previous systems \cite{fitzgerald2015semantic,li2019dependency} were evaluated on the original CoNLL-05, CoNLL-09 datasets and the performance is only a reference for rough comparison.}\label{tab:joint-results}
	\begin{tabular}{lcccccccccccc}  
		\toprule  
		\multirow{2}{*}{\textit{Gold predicates}} &\multicolumn{3}{c}{Span WSJ}&\multicolumn{3}{c}{Span Brown}&\multicolumn{3}{c}{Dep. WSJ}&\multicolumn{3}{c}{Dep. Brown}\\
		\cmidrule(lr){2-4} \cmidrule(lr){5-7} \cmidrule(lr){8-10}  \cmidrule(lr){11-13}
		&P&R&F$_1$&P&R&F$_1$&P&R&F$_1$&P&R&F$_1$\\
		\midrule  
		\textit{wo/ PLM} & & & & & & & & & & & & \\
		\citet{fitzgerald2015semantic} &82.3&76.8&79.4&73.8&68.8&71.2 & - & - & 87.3  & -  & - & 75.2 \\
		\citet{li2019dependency}  & - & - & 83.0 & - & - & - & - & - & 85.1 & - & - & - \\
		\textbf{Baseline} & 82.9 & 83.9 & 83.4 & 72.9 & 69.8 & 71.3 & 85.9 & 84.7 & 85.3 & 76.2 & 75.0 & 75.6 \\
		\textbf{Baseline-joint} & 83.9  & 83.5 & 83.7 &  73.5 & 69.1 &  71.2 &  86.7  &  86.9 & 86.8 & 75.2 & 75.4  &  75.3 \\
		\quad\textbf{+Predicted Syntax} & 85.0  & 86.0  & 85.5  &  74.0 & 70.8 & 72.3 &  87.8  & 88.0  &  87.9 & 77.0  & 76.8 & 76.9 \\
		\quad\textbf{+Gold Syntax} & 88.0 & 86.4 & 87.2  & 76.6  & 71.2  & 73.8  & 89.1 & 88.7 &  88.9 &  78.9 & 76.7 & 77.8 \\
		\midrule
		\textit{w/ ELMo} & & & & & & & & & & & & \\
		\citet{li2019dependency}  & 87.9 & 87.5 & 87.7& 80.6& 80.4 & 80.5 &89.6&91.2&90.4&81.7 &81.4&81.5 \\
		\textbf{Baseline} & 88.1 & 86.9 & 87.5 & 81.1 & 80.1 & 80.6 &  90.6 & 90.2 & 90.4 & 80.9 & 82.3 & 81.6 \\
		\textbf{Baseline-joint} & 88.2 & 87.6 & 87.9 & 81.0 & 80.8 & 80.9 &  90.0 & 91.2 & 90.6 & 81.7 & 81.5 & 81.6 \\
		\quad\textbf{+Predicted Syntax} & 88.5 & 88.1 & 88.3 & 81.3 & 81.1 & 81.2 &  90.5 & 92.1 & 91.3 & 81.7 & 81.9 & 81.8 \\
		\quad\textbf{+Gold Syntax} & 89.6 & 90.1 & 89.8 & 82.4 & 82.6 & 82.5 &  90.8 & 93.5 & 92.2 & 82.0 & 83.4 & 82.7 \\
		\midrule
		\textit{w/ BERT} & & & & & & & & & & & & \\
		\textbf{Baseline} & 88.3 & 88.9 & 88.6 & 81.1 & 81.9 & 81.5 & 91.1 & 91.5 & 91.3 & 82.6 & 81.8 & 82.2 \\
		\textbf{Baseline-joint} & 87.9  & 89.7 & 88.8 & 81.4 & 81.6 & 81.5 & 91.2 & 91.4 & 91.3 & 82.8 &  82.2 & 82.5 \\
		\quad\textbf{+Predicted Syntax} & 88.9 & 89.1 & \textbf{89.0} & 81.6 & 82.0 & \textbf{81.8} & 91.4  & 91.4 & \textbf{91.4} & 82.4 & 82.8 & \textbf{82.6} \\
		\quad\textbf{+Gold Syntax} & 90.2 & 91.8 & 91.0 & 83.2 & 84.0 & 83.6 & 92.4 & 93.0 & 92.7 & 82.8 & 84.8 & 83.8 \\
		\bottomrule  
	\end{tabular}
\end{table*}

\subsection{Main Results}

We report the evaluation results with the \emph{gold predicates} setup on the ConSD test set in Table \ref{tab:joint-results} and also include the results of the latest works on the CoNLL-05 and CoNLL-09 benchmarks. Comparing the results of \textbf{Baseline} and \cite{li2019dependency}, both use the same model architecture and training mode, we found that the evaluation scores on the span and dependency test sets of ConSD approximate those on  CoNLL-05 and CoNLL-09, respectively. Therefore, we can make a rough comparison between the results of our model on ConSD and models evaluated  on CoNLL-05 and CoNLL-09 benchmarks.

Taking advantage of the fact that we have integrated span and dependency styles in the dataset, in contrast to previous methods, we used a single model (instead of two independent models) to optimize the joint SRL objectives. When comparing our joint model \textbf{Baseline-joint} with the independently trained system \textbf{Baseline}, though the models have similar structure, the performance of the \textbf{Baseline-joint} system is superior to \textbf{Baseline}. This demonstrates the effectiveness of the joint optimization across different styles.

With the inclusion of pre-trained language model features (ELMo, BERT), joint training still shows some performance improvements over the strong baselines. Since a joint model can output dependency and span SRL annotations at the same time, our training and inference time are nearly halved compared to using two independently optimized SRL models.

Using our new proposed ConSD dataset and the novel joint baseline, we also reported the effect of syntax information on SRL. As shown in the table, the introduction of syntactic information brings extra improvements to the baseline model, which demonstrates the role of syntax information in SRL and the completeness of our dataset, as it does in fact include useful syntax information. Additionally, we also demonstrate that the proposed new syntax-aided method firmly and uniformly boosts both styles of SRL when using predicted syntax. Although gold syntax is not available in practical applications, it can still be used to draw some conclusions. The group using gold syntax in the experiment achieved a significant increase in recall, indicating that the boundary and head position indicator embeddings are useful for identifying arguments.

\begin{figure}[h]
	\begin{minipage}{0.49\linewidth}
		\centering
		\small
		\setlength{\tabcolsep}{4pt}
		\makeatletter\def\@captype{table}\makeatother\caption{The span SRL results on CoNLL-05 and ConSD test sets with the \emph{end-to-end} setup. $^\dag$ indicates that pre-trained language model ELMo is used in that systems.}\label{tab:end-to-end-results}
		\begin{tabular}{lcccccc}
			\toprule  
			\multirow{2}{*}{\textit{End-to-end}}&\multicolumn{3}{c}{WSJ}&\multicolumn{3}{c}{Brown}\\  
			\cmidrule(lr){2-4} \cmidrule(lr){5-7}
			&P&R&F$_1$&P&R&F$_1$ \\  
			\midrule
			\multicolumn{7}{c}{\bf CoNLL-05}\\
			\citet{he-acl2017}  &80.2&82.3&81.2&67.6&69.6&68.5\\
			\citet{he2018jointly}$^\dag$  &84.8&87.2&86.0&73.9&78.4 &76.1\\  
			\citet{Strubell2018}$^\dag$ &87.1 &86.7 &86.9& 79.0 &77.5 &78.3 \\
			\citet{li2019dependency}$^\dag$ &85.2&87.5&86.3&74.7& 78.1&76.4\\
			\midrule
			\multicolumn{7}{c}{\bf ConSD}\\
			\textit{w/ ELMo} & & & & & & \\
			\textbf{Baseline} & 86.2 & 85.8 & 86.0 & 77.1 & 75.9 & 76.5 \\
			\textbf{Baseline-joint} & 86.7 & 86.1 & 86.4 & 75.3 & 78.1 & 76.7\\
			\quad\textbf{+Predicted Syntax} & 86.7 & 86.7 & 86.7 & 76.4 & 78.2 & 77.3\\
			\quad\textbf{+Gold Syntax} & 88.6 & 88.0 & 88.3 & 78.0 & 78.2 & 78.1 \\
			\hdashline
			\textit{w/ BERT}  & & & & & & \\
			\textbf{Baseline} & 87.8 & 86.4 & 87.1 & 80.9 &  78.5 & 79.7\\
			\textbf{Baseline-joint} & 88.1 & 86.3 & 87.2 & 79.7 & 79.5 & 79.6 \\
			\quad\textbf{+Predicted Syntax} & 88.1 & 87.5 & \textbf{87.8} & 79.8  & 80.2 &  \textbf{80.0} \\
			\quad\textbf{+Gold Syntax} & 88.4 & 89.6 & 89.0 & 80.3 & 83.9 & 82.1 \\
			\bottomrule
		\end{tabular}
		
	\end{minipage}
	\hfill
	\begin{minipage}{0.49\linewidth}
		\centering
		\small
		\setlength{\tabcolsep}{4pt}
		\makeatletter\def\@captype{table}\makeatother\caption{The dependency SRL results on CoNLL-09 and ConSD test sets with the \emph{end-to-end} setup. $^\dag$ indicates that pre-trained language model ELMo is used in that systems.}\label{tab:end-to-end-results1}
		\begin{tabular}{lcccccc}
			\toprule  
			\multirow{2}{*}{\textit{End-to-end}}&\multicolumn{3}{c}{WSJ}&\multicolumn{3}{c}{Brown}\\  
			\cmidrule(lr){2-4} \cmidrule(lr){5-7}
			&P&R&F$_1$&P&R&F$_1$ \\  
			\midrule
			\multicolumn{7}{c}{\bf CoNLL-09}\\
			\citet{he:2018Syntax}$^\dag$  &83.9&82.7&83.3&$-$&$-$&$-$\\
			\citet{cai2018full} &84.7&85.2&85.0&$-$&$-$&72.5\\
			\citet{li2019dependency}$^\dag$  &84.5&86.1&85.3&74.6&73.8&74.2 \\  
			\midrule
			\multicolumn{7}{c}{\bf ConSD}\\
			\textit{w/ ELMo} & & & & & & \\
			\textbf{Baseline} & 85.7 & 84.7 & 85.2 & 73.8 & 74.2 & 74.0 \\
			\textbf{Baseline-joint} & 85.0 & 86.2 & 85.6 & 74.6 & 74.0 & 74.3\\
			\quad\textbf{+Predicted Syntax} & 86.1 & 85.9 & 86.0 & 75.0 & 74.8 & 74.9\\
			\quad\textbf{+Gold Syntax} & 88.6 & 88.9 & 88.7 & 75.7 & 75.3 & 75.5\\
			\hdashline
			\textit{w/ BERT} & & &   & & & \\
			\textbf{Baseline} & 87.2 & 86.6 & 86.9 & 75.6 & 76.4 & 76.0 \\
			\textbf{Baseline-joint} & 87.9 & 86.3 & 87.1 & 77.1 & 74.9 & 76.0 \\
			\quad\textbf{+Predicted Syntax} & 87.8 & 88.0 & \textbf{87.9} & 77.1  & 75.5 & \textbf{76.3} \\
			\quad\textbf{+Gold Syntax} & 88.6 & 90.4 & 89.5  & 76.0  & 80.5  & 78.2 \\
			\bottomrule
		\end{tabular}
		
	\end{minipage}
\end{figure}

The \textit{end-to-end} setup is a more difficult SRL setting in which the predicate is not pre-specified. We also evaluated the model under this setting on the ConSD dataset, as shown in Table \ref{tab:end-to-end-results} and \ref{tab:end-to-end-results1}. From the results, joint optimization and syntactic enhancement brought about an improvement consistent with that in the \textit{gold predicate} setup.

\section{Ablation}

\subsection{Performance Comparison}

Due to the changes in the dataset (both the training set and the test set), the results that we achieved on the ConSD dataset were not directly comparable to those trained and evaluated on the original CoNLL-05 and CoNLL-09 datasets. Since our ConSD datasets still follow the annotation rules of span and dependency style SRL, the models trained on ConSD dataset can still be evaluated on the original CoNLL-05, CoNLL-09 test sets for direct comparison of results.
The results of this cross-evaluation are reported in Table \ref{tab:original_new}. Since the CoNLL-05 test set does not contain nominal predicates, we remove all non-verb predicates and their arguments in the prediction according to the POS tags of the predicates to make the results roughly comparable.

In span style, looking the performance of the same model on the CoNLL-05 and ConSD test sets, the $F_1$ score on the CoNLL-05 test set was lower than that on the ConSD test set, indicating that labeling roles for nominal predicates was easier than doing so for verbal predicates. The model focused less on the verbal predicates because of  the mixed training of nominal predicates and verbal predicates, so the results on verbal predicates were lower than \citet{li2019dependency}. This suggests that we can adopt different models according to the types of predicates to improve the overall performance of SRL. In addition, the granularity of some words in ConSD is inconsistent with that of their counterparts in CoNLL-05, which is another reason for decreased performance. In the dependency style, the results of the CoNLL-09 and ConSD test sets are virtually the same for a single model, indicating that the difference between the ConSD test set and the CoNLL-09 test set is very small.

\begin{table}[h]
	\caption{End-to-end SRL performance evaluated on CoNLL-05, CoNLL-09, and ConSD test sets. The models in \cite{li2019dependency} are trained separately on the CoNLL-05, CoNLL-09 training set tests and then tested in the corresponding test sets. Our models are trained on the training set of ConSD, and then report the results on the test set of CoNLL-05, CoNLL-09 and ConSD.}\label{tab:original_new}
	\begin{tabular}{lcccccccccccc}
		\toprule  
		\multirow{3}{*}{\textit{End-to-end w/ ELMo}}&\multicolumn{6}{c}{Span}&\multicolumn{6}{c}{Dependency}\\  
		\cmidrule(lr){2-7} \cmidrule(lr){8-13} & \multicolumn{3}{c}{CoNLL-05} & \multicolumn{3}{c}{ConSD} & \multicolumn{3}{c}{CoNLL-09} & \multicolumn{3}{c}{ConSD} \\
		\cmidrule(lr){2-4} \cmidrule(lr){5-7} \cmidrule(lr){8-10} \cmidrule(lr){11-13} &P&R&F$_1$&P&R&F$_1$ &P&R&F$_1$&P&R&F$_1$\\  
		\midrule
		\citet{li2019dependency} &  85.2&87.5&86.3 & $-$ & $-$ & $-$ & 84.5&86.1&85.3 & $-$ & $-$ & $-$\\
		\textbf{Baseline} & 84.7 & 86.3 & 85.5 & 86.2 & 85.8 & 86.0 & 85.6 & 84.6 & 85.1 & 85.7 & 84.7 & 85.2 \\
		\textbf{Baseline-joint} & 86.4 & 85.6 & 86.0 & 86.7 & 86.1 & 86.4 & 85.0 & 86.2 & 85.6 &  85.0 & 86.2 & 85.6 \\
		\textbf{Full Model} & 85.5 & 87.1 & 86.3 & 86.7 & 86.7 & 86.7 & 86.0 & 85.9 & 86.0 & 86.1 & 85.9 & 86.0 \\ 
		\bottomrule
	\end{tabular}
\end{table}

\subsection{Syntax Role}

In order to assess the effect of our two proposed syntax-aided methods on span-based SRL and dependency-based SRL, respectively, we conducted an ablation study on the WSJ test set with the \emph{end-to-end} setup using predicted syntax and BERT pre-trained language model (LM) features to explore how the different syntax (constituency and dependency) impact our model. As the results show in Table \ref{tab:ablation}, in our joint model, \textbf{CSA} with constituent syntax has a greater impact than \textbf{DSA} with dependency syntax, as \textbf{CSA} affects the argument span decision, and the \textbf{DSA} affects the head position and also contributes to the representation of spans. This shows that our two syntactic enhancements are effective.

\begin{table}[h]
	\caption{The different contributions of syntax for span-based and dependency-based SRL with the \emph{end-to-end} setup using predicted syntax and BERT pre-trained LM features.}\label{tab:ablation}
	\begin{tabular}{lcccccc}
		\toprule  
		\multirow{2}{*}{End-to-end}&\multicolumn{3}{c}{Span SRL}&\multicolumn{3}{c}{Depedency SRL}\\  
		\cmidrule(lr){2-4} \cmidrule(lr){5-7}
		&P&R&F$_1$&P&R&F$_1$ \\  
		\midrule
		\textbf{Full Model} & 88.1 & 87.5 & 87.8 & 87.8 & 88.0 & 87.9  \\
		\quad\textbf{-DSA} & 88.0 & 87.4 & 87.7  & 88.1  & 86.5 & 87.3 \\
		\quad\textbf{-CSA} & 87.9 & 86.7 & 87.3  & 88.3  & 86.5  & 87.4 \\
		\quad\textbf{-Both} & 88.1 & 86.3 & 87.2 & 87.9 & 86.3 & 87.1 \\
		\bottomrule
	\end{tabular}
\end{table}

\section{Related Work}
In the early work of semantic role labeling, most researchers focused on feature engineering \cite{xue2004,pradhan2005,punyakanok2008importance}. 
Putting syntax aside has sparked much research interest since \citet{zhou-xu2015} employed deep BiLSTMs for span SRL. A series of neural SRL models without syntactic inputs were proposed. \citet{marcheggiani2017} applied a simple LSTM model with effective word representation, achieving encouraging results on English, Chinese, Czech,  and Spanish. \citet{cai2018full} built a full end-to-end SRL model with biaffine attention and provided strong performance on English and Chinese. \citet{li2019dependency} also proposed an end-to-end model for both dependency and span SRL with a unified argument representation, obtaining favorable results on English. 

Despite the success of syntax-agnostic SRL models, more recent work features attempts to further improve performance by integrating syntactic information. \citet{marcheggianiEMNLP2017} used graph convolutional network to encode syntax into dependency SRL. \citet{he:2018Syntax}  proposed an extended $k$-order argument pruning algorithm based on syntactic tree and boosted SRL performance. 
\citet{li2018emnlp} presented a unified neural framework to provide multiple methods for syntactic integration. 
Our method is closely related to that of \cite{he:2018Syntax}, which was designed to prune as many unlikely arguments as possible. 
While the above models only considered dependency syntax, an alternative syntax representation available for potential use is constituency syntax. \citet{he-acl2017} treated constituency syntax trees as indications of argument boundaries; however, they only used this boundary information during sequence decoding. 
In their $A*$ search algorithm, the sequences with arguments that missed or crossed the constituent boundaries were penalized.
All of the above models only considered either the dependency or the span SRL formalism, though the latest work \cite{li2019dependency} uses two separate but similarly designed models to handle the two formalisms. Rather than using two models, we instead propose a joint SRL formalism for both dependency and span style.

\section{Conclusion}
We build a new cross-style SRL dataset, ConSD, that, for the first time, jointly considers dependency and span SRL formalisms with a linguistic motivation. Furthermore, with a new cross-style joint syntax-agnostic model and syntax-aided features, we explored diverse settings on the new proposed dataset, which verifies that the proposed cross-style SRL convention is also helpful for computational purposes. 

\bibliographystyle{ACM-Reference-Format}
\bibliography{reference}


\end{document}